\documentclass{article}
\usepackage{spconf,amsmath,graphicx}
\pdfoutput=1

\usepackage{enumitem}
\setlist{nosep, leftmargin=14pt}

\usepackage{trimclip}
\usepackage{stmaryrd}
\usepackage{amsfonts}

\newcommand{\cropbar}[1]{\clipbox*{0pt -.4ex {\width} .5ex}{#1}}
\DeclareMathOperator*{\argmax}{argmax}

\title{Utilizing uncertainty estimation in deep learning segmentation of fluorescence microscopy images with missing markers}
\name{Alvaro Gomariz$^{1, 2}$, Raphael Egli$^{1}$, Tiziano Portenier$^{1}$, C\'esar Nombela-Arrieta$^{2}$, Orcun Goksel$^{1, 3}$}
\address{
$^{1}$ Computer-assisted Applications in Medicine, ETH Zurich, Switzerland \\
$^{2}$ Dept.\ of Medical Oncology \& Hematology, University Hospital \& University of Zurich, Switzerland \\
$^{3}$ Department of Information Technology, Uppsala University, Sweden
}

\begin{document}

\maketitle
\begin{abstract}
Fluorescence microscopy images contain several channels, each indicating a marker staining the sample. 
Since many different marker combinations are utilized in practice, it has been challenging to apply deep learning based segmentation models, which expect a predefined channel combination for all training samples as well as at inference for future application.
Recent work circumvents this problem using a modality attention approach to be effective across any possible marker combination. 
However, for combinations that do not exist in a labeled training dataset, one cannot have any estimation of potential segmentation quality if that combination is encountered during inference.
Without this, not only one lacks quality assurance but one also does not know where to put any additional imaging and labeling effort.
We herein propose a method to estimate segmentation quality on unlabeled images by 
($i$)~estimating both aleatoric and epistemic uncertainties of convolutional neural networks for image segmentation, and 
($ii$)~training a Random Forest model for the interpretation of uncertainty features via regression to their corresponding segmentation metrics.
Additionally, we demonstrate that including these uncertainty measures during training can provide an improvement on segmentation performance.

\end{abstract}

\section{Introduction}
\label{sec:intro}
Recent progress in Convolutional Neural Networks (CNNs) for image segmentation has enabled many possibilities in biomedical tissues analysis using fluorescence microscopy (FM) \cite{unet_falk2019u}, which images fluorescent dyes (called \emph{markers}) that target different cells or anatomical networks in biological specimens \cite{gomariz2020imaging}. 
Different markers are combined and registered as image channels, which are for our purposes equivalent to modalities in other imaging techniques such as with different sequences in MRI. 
However, the maximum number of markers in a FM sample is limited due to their spectral overlap, thus requiring different combinations for each biological study. 
Furthermore, the lack of immunostaining consistency due to sample preparation difficulties, together with limited sample availability, often leads to datasets of images with missing markers, which have been challenging for traditional CNNs.
To address this, a modality sampling and attention approach named \emph{Marker Sampling \& Marker Excite} (\mbox{\emph{MS-ME}}) was proposed  in \cite{gomariz2020modality}, which allows for training and inference with a single model on datasets with heterogeneous marker combinations.
Although \mbox{\emph{MS-ME}} permits predictions on marker combinations unseen during training, one does not know in advance what segmentation quality to expect for such unseen combinations without any existing labeled data. 
Such quality estimators would be very valuable not only in predicting potential future performance, but also in deciding where to invest additional imaging and labeling effort.

Uncertainty estimation in deep neural networks with different Bayesian approximations was shown to improve model predictions, either by explaining this within the loss function or by aggregating predictions from ensembles~\cite{kendall2017uncertainties}.
Such estimations have been successfully applied to medical image segmentation~\cite{epistemicprobs_hiasa2019automated}.
\emph{Epistemic} uncertainty accounts for lack of confidence in the parameters of a model, i.e.\ uncertainty that can be reduced with additional labeled data. This can be estimated using so-called Dropout layers~\cite{srivastava2014dropout} both at training and inference time, known as Monte Carlo (MC) dropouts~\cite{gal2015dropout,gal2015cnn}.
\emph{Aleatoric} uncertainty captures the inherent noise in the observations, i.e.\ uncertainty that cannot be reduced with additional data; and it can be estimated with the inclusion of a stochastic loss as proposed in~\cite{kendall2017uncertainties}.
We herein study the use of the above uncertainty estimation tools to design a method for the estimation of segmentation quality in the above-described problem setting of heterogeneous FM marker combinations. 

\section{Methods}
\begin{figure}[htb]
    \centering
    \includegraphics[width=1.\linewidth]{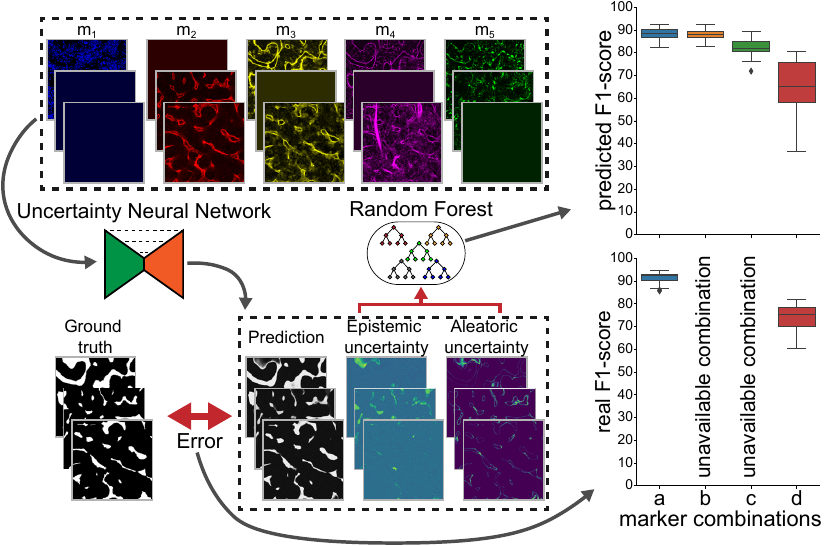}
    \caption{Illustration of our proposed framework for predicting segmentation quality with missing markers. 
    }
    \label{fig:illustration}
\end{figure}

As illustrated in Fig.\  \ref{fig:illustration}, we extend \mbox{\emph{MS-ME}} to also estimate aleatoric and epistemic uncertainties, using the summary statistics of which we train a regression Random Forest (RF) to predict quantitative segmentation outcomes.

\subsection{Learning from images with missing markers}
FM images are formed by different combinations of markers $m_G$, with $G \subseteq \{1, \dots, K\}$, and $K$ the number of possible markers represented as channels in an image $x \in \mathbb{R}^{h \times w \times |G|}$.
We denote the combination as the successive indexes of the markers it contains, e.g.\ $m_{24}$ is a combination of markers 2 and 4. 
The challenge of different marker combinations in FM images for both in training and testing was addressed in \cite{gomariz2020modality} using \mbox{\emph{MS-ME}}.
Marker sampling (MS) refers to MC dropout of modalities at training time, to make testing generalizable to different availability of markers.  
For Marker Excite (ME), a feature-wise attention module that has 2 fully-connected layers and a one-hot encoded input of available modalities for the sample is added at different layers of a UNet~\cite{unet_falk2019u}.  
For a detailed network structure and implementation details, see~\cite{gomariz2020modality}.

\subsection{Uncertainty estimation in CNN-based segmentation}
Different uncertainties are estimated following the frameworks proposed in \cite{kendall2017uncertainties, epistemicprobs_hiasa2019automated} and included within a \mbox{\emph{MS-ME}} model $f$ that is applied on an input image $x$ to predict its corresponding segmentation $\hat{y} = \textit{softmax}(z)$, where $z$ are the logits resulting from the model: $ z = f(x)$.

To estimate epistemic uncertainty $u_\mathrm{e}$ of $x$, MC Dropout is employed at different layers of a CNN $f_{\mathrm{e}(p)}$ with probability $p$ both at training and inference. 
Since the output of the network is stochastic, $\hat{y}$ and $u_\mathrm{e}$ are estimated respectively as the mean and standard deviation (SD) of $T$ samples from $f_{\mathrm{e}(p)}(x)$.
When explicitly stated, we add MC Dropout only after the last layer as proposed in~\cite{gal2015dropout}.  

Aleatoric uncertainty $u_\mathrm{a}$ is calculated by explicitly adding a predictive variance $u_\mathrm{a}$ to our model output, i.e.\ $
    [z,\, u_a] = f_\mathrm{a}(x)$\,.
This model $f_\mathrm{a}$ is trained with a stochastic cross-entropy loss that adds a noise component $\epsilon  \sim \mathcal{N}(0,\, I)$ to multiple ($T$) model predictions, so that the loss evaluates their mean: 
\begin{equation*}\label{eq:aleatoric_training}
    \hat{y}^\text{training} = \frac{1}{T} \sum_t^T \textit{softmax}(z + u_a \epsilon_t)
\end{equation*}
At inference, both $\hat{y}$  and $u_a$ can be obtained without sampling. 

Both techniques are simultaneously included in a single \emph{combined} model $f_{\mathrm{e+a}(p)}$ that is employed to separately estimate both $u_e$ and $u_a$. 
The number of prediction samples is set to $T=50$ in our experiments.

\subsection{Predicting segmentation quality from uncertainty}
Different uncertainty measures only provide visual cues about prediction errors.
We herein, however, aim to obtain a quantitative estimation of the segmentation quality $q$.
To this end, we propose different regression models $g$ to obtain quality predictions $\hat{q} = g(u)$ from uncertainties $u$ obtained from $f(x)$ as described above.
We train $g$ on $u$ extracted from all possible marker combinations within the validation set of the segmentation task, and compare $\hat{q}$ with the ground truth $q$ extracted from their corresponding segmentations. For comparison, herein we use the $F1$-score for binary classification, i.e.:
\begin{equation*}
    q = \frac{2 \left|y \cap \argmax(\hat{y})\right|}{|y| + |\argmax(\hat{y})|}\ .
\end{equation*}
We subsequently evaluate $\hat{q}$ on the test set using a Root Mean Squared Error (RMSE) metric with respect to $q$ across all possible marker combinations.
We employ 4-fold cross-validation on the same data split as the segmentation task.

A first approach tested is to train an additional CNN $g_\mathrm{CNN}$ to predict $\hat{q} = g_\mathrm{CNN}(\{u_e, u_a\})$. 
We design a simple regression architecture with a very small number of parameters to avoid overfitting to the limited size of the validation set.
By denoting 2D convolutional layers with $l$ nodes and $3\times3$ kernels as $\text{C}(l)$, 2D max pooling layers with $3\times3$ kernel as \text{MP}, and fully connected layers with $l$ nodes as $\text{FC}(l)$, we use the following CNN:
$\text{C}(4) \shortrightarrow \! \text{MP} \!
\shortrightarrow\!\text{C}(8) \shortrightarrow \! \text{MP} \!
\shortrightarrow\!\text{C}(16) \shortrightarrow \! \text{MP} \!
\shortrightarrow\!\text{C}(32) \shortrightarrow \! \text{MP} \!
\shortrightarrow\!\text{C}(64) \shortrightarrow \! \text{FC}(128) \!
\shortrightarrow \! \text{FC}(1)$.
We add a ReLu activation after every convolutional or fully connected layer, and use a batch size of 2. 
We train for 100 epochs with $L2$ loss and Adam optimizer \cite{kingma2014adam} with a learning rate of $10^{-3}$. 

Since different uncertainty maps may contain error-related information that qualitatively correlate with $q$, we hypothesize that traditional machine-learning models that have far lower degrees-of-freedom by utilizing hand-crafted globally informative features are potentially more robust to overfitting to the limited data available for this task. 
To this end, we alternatively train a RF model $g_\mathrm{RF}$ with 128 trees, using mean squared error as split criterion. 
As features we use: uncertainty map percentiles (99 values from 1\textsuperscript{st} to 99\textsuperscript{th}), its cumulative histogram (in 13 bins from values 0.05 to 0.65), its first four statistical moments, and a one-hot vector indicating which among all possible marker combinations is used. 
We compare three approaches:  $g_\mathrm{RF}(u_e)$ with only epistemic features, $g_\mathrm{RF}(u_a)$ with only aleatoric, and $g_\mathrm{RF}(\{u_e, u_a\})$ with both.

\section{Results and Discussion}
\subsection{Dataset and use of markers}
We employ the FM dataset of bone marrow vasculature described in \cite{gomariz2018quantitative} with the experimental settings in \cite{gomariz2020imaging}.
The dataset contains 8 samples decomposed into different 2D patches (a total of 230), each with $K=5$ markers. 
We use the annotated class \emph{sinusoids}, which amounts to 11.41\% of the pixels, the rest being considered background. 
The samples are divided into 5 for training, 1 for validation, and 2 for testing. 
Since the segmentation quality vastly differs across marker combinations, we evaluate a relative segmentation $F1$-score with respect to a reference model on the test set. This score compares the pair-wise $F1$-scores across 4 cross-validation steps and all possible combinations (31) of the 5 available markers. 
Since it is not feasible to study all possible marker combinations in the training set, 7 training scenarios are proposed.
In the first one, all 5 markers are available in all 5 training samples. 
In the rest, test cases \#1 to \#6 given in~\cite{gomariz2020imaging}, where different markers are artificially ablated on each of the samples, are adopted herein for comparability with the baseline. 

\subsection{Utilizing uncertainties in segmentation}
\begin{figure}[htb]
    \centering
    \includegraphics[width=1.\linewidth]{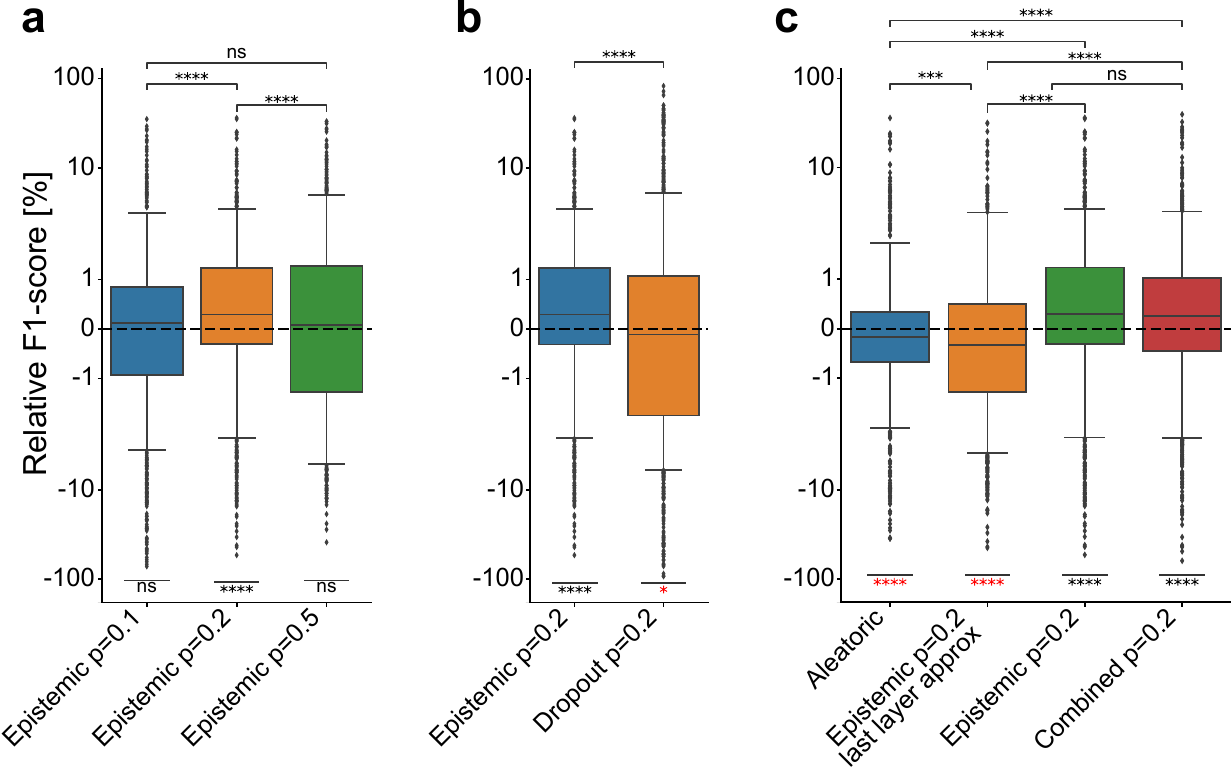}
    \caption{
    Segmentation performance of different proposed models $f$ evaluated w.r.t.\ baseline \mbox{\emph{MS-ME}}
    \ (n=868). 
    \textbf{(a)}~Effect of different $p$ in $f_\mathrm{e}(p)$. \textbf{(b)}~Comparison with conventional Dropout at training. 
    \textbf{(c)}~Comparison of the proposed uncertainty-based models.
    Symmetrical logarithmic scale (linear between -2 and 2, logarithmic elsewhere) is used. 
    Significance is shown between different models  (\protect\cropbar{$|$}\!-----\!\protect\cropbar{$|$}), or w.r.t.\ the reference \emph{MS-ME} (-----), in red when a proposed model performs worse.
    }
    \label{fig:performance_all}
\end{figure}

We herein study how the segmentation $F1$-score is affected in our missing markers framework when accounting for the presented uncertainty methods in the \mbox{\emph{MS-ME}} architecture. 
We compare the performance to a reference \mbox{\emph{MS-ME}} without uncertainty estimation across all 31 possible marker combinations, 4 cross-validation steps, and 7 missing marker scenarios. 
The results in Fig.\ \ref{fig:performance_all}(a) show that in the estimation of $u_e$, it is best to use MC dropout in all convolutional layers with $p=0.2$ ($f_\mathrm{e}(p=0.2)$), which is also superior to baseline \mbox{\emph{MS-ME}}.
We assess that the $F1$-score superiority is not merely due to the use of Dropout by comparing it in Fig.\ \ref{fig:performance_all}(b) to a \mbox{\emph{MS-ME}} model with standard Dropout layers (not sampling at inference) with the same probability, which is significantly worse than \mbox{\emph{MS-ME}} and $f_\mathrm{e}(p=0.2)$.

Using the stochastic loss for the estimation of $u_a$ with $f_\mathrm{a}$ leads to an inferior $F1$-score as seen  in Fig.\ \ref{fig:performance_all}(c). 
But as discussed in the next part, such information is desirable regardless of the negative results.
Thus, we employ $f_\mathrm{e+a}(p=0.2)$ that allows for prediction of both $u_e$ and $u_e$ while providing the best $F1$-score together with $f_\mathrm{e}(p=0.2)$.

\subsection{Predicting segmentation quality for the unseen}
In addition to the superior $F1$-score achieved with the use of $f_\mathrm{e+a}(p=0.2)$, the simultaneous estimation of $u_e$ and $u_a$ allows to visually inspect potential mistakes in the CNN prediction.
Here, we use such uncertainty maps to estimate the $F1$-score $q$ of segmented images.
We evaluate our proposed methods on the training setting denominated case \#6, for which patches have marker combinations $m_{135}$, $m_{124}$, $m_{35}$, $m_{23}$, or $m_{45}$ depending on which of the 5 training samples they belong to. 
This setting contains a variety of markers for each sample that depicts scenarios usually found in practise, and it allows to evaluate the predicted $q$ in markers not available for any of its training samples.

\begin{figure}
    \centering
    \includegraphics[width=1\linewidth]{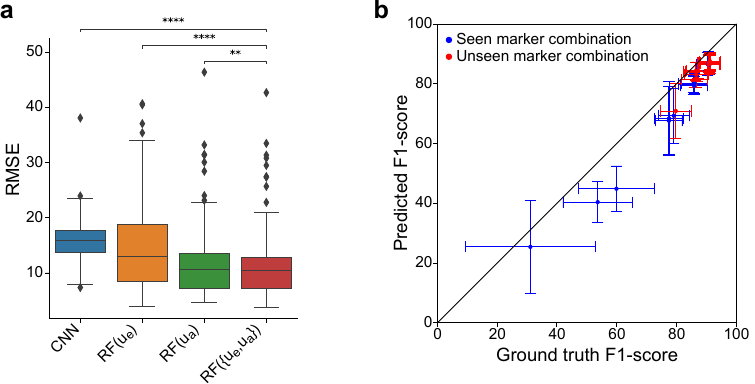}
    \caption{
    Prediction of $F1$-score from uncertainty maps on the test set. 
    \textbf{(a)}~RMSE of different models (n=124). 
    \textbf{(b)}~Comparison of predicted $F1$-score using $g_\mathrm{RF}(\{u_e, u_a\})$ to ground truth for each of the 31 marker combinations shown as point (mean) and bars (SD)\ (n=4). 
    }
    \label{fig:fscoreprediction}
\end{figure}

We show in Fig.\ \ref{fig:fscoreprediction}(a) the RMSE results for the $F1$-score prediction with the methods presented above. 
Using $g_\mathrm{CNN}$ leads to worse RMSE results than any of the $g_\mathrm{RF}$ methods, which can be explained by the tendency of CNNs to overfit in such a small training set, and that the simplicity of the task asks for the efficient use of predefined explanatory features where RF methods excel.
$g_\mathrm{RF}(\{u_e, u_a\})$ is seen to be superior in RMSE to both $g_\mathrm{RF}(u_e)$ and $g_\mathrm{RF}(u_a)$.
This observation can be attributed to the fact that $u_e$ and $u_a$ capture information about distinct segmentation errors that both help in the regression of a more accurate $F1$-score value. 
Although further improvements could be achieved with deep ensembles, which have been reported to estimate more accurate uncertainties than MC dropout~\cite{fort2020deep}, they are computationally very expensive, which would annul a main advantage of our framework of training a single model. 

We subsequently employ $g_\mathrm{RF}(\{u_e, u_a\})$ to estimate $F1$-score for both seen and unseen marker combinations in our training scenario, which is observed in Fig.\ \ref{fig:fscoreprediction}(b) to closely follow ($R^2$$=$$0.98$) the ground-truth predictions on average across cross-validation steps.
Thus, despite potentially high standard deviations (and high RMSE) per individual patch predictions, we can still make accurate overall predictions about the expected quality of a marker combination, including those that are unseen.

\begin{figure}
    \centering
    \includegraphics[width=1.0\linewidth]{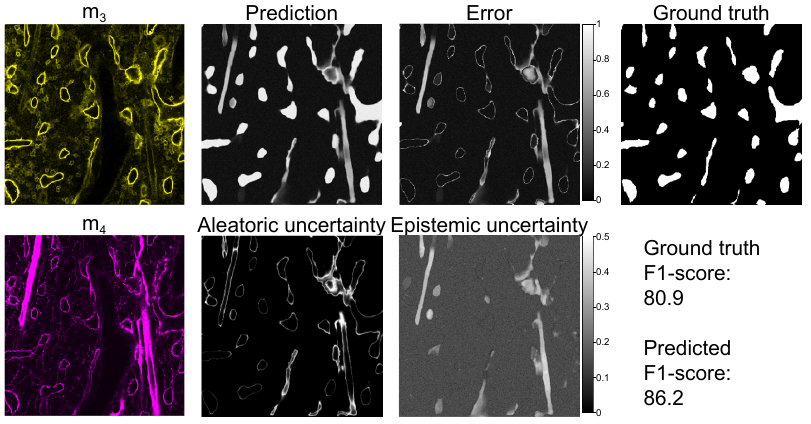}
    \caption{
    Visual example of a prediction and their uncertainties for the test set marker combination $m_{34}$ in training case \#6.
    The error image shows both false positives and negatives. 
    The $F1$-score is predicted using $g_\mathrm{RF}(\{u_e, u_a\})$.
    }
    \label{fig:visualization}
\end{figure}

We also visually assess the individual effects of $u_e$ and $u_a$ in Fig.\ \ref{fig:visualization}.
$u_a$~is observed to focus on the edges of the vessels, where discrepancies in annotations often exist and thus fit the definition of aleatoric uncertainty of capturing noise inherent in the observations. 
Meanwhile, $u_e$~captures mistakenly segmented vessels which appear in $m_4$ but are not sinusoids. 
Such errors are related to lack of information in the model, and may be corrected by adding labeled training data containing patterns similar to those highlighted by $u_e$.

\section{Conclusion}
With the proposed framework for inspection of segmentation uncertainties with missing FM markers, the advantages are twofold: 
First, accounting for epistemic and aleatoric uncertainties simultaneously produces segmentation results superior to the state-of-the-art \emph{MS-ME} model.
Second, with a uncertainty feature based regressor, we can estimate segmentation quality for any possible marker combination, whether it was seen during training or not. 
Thus, in practise we can quantitatively evaluate how suitable a trained model is for future samples stained with a previously unseen marker combination, i.e.\ to discard them or preferentially annotate them, e.g., in an active learning framework.
Furthermore, comparison of different uncertainties may indicate whether to focus more on labeled data or on improving image acquisition quality. 
Note that the proposed methods are applicable also to other multi-modality frameworks, such as MRI.

\ninept
\section{Compliance with Ethical Standards}
\label{sec:ethics}
This research was conducted on a dataset described in \cite{gomariz2018quantitative, gomariz2020modality}, where the ethical compliance for sample preparation was described.

\section{Acknowledgments}
\label{sec:acknowledgments}
Funding was provided by Hasler Foundation, Swiss National Science Foundation (SNSF), Swiss Cancer Research Foundation, and Julius-M{\"u}ller Foundation.

\bibliographystyle{IEEEbib}
\bibliography{refs}

\end{document}